\documentclass[11pt,letterpaper]{article}
\usepackage{emnlp2016}
\usepackage{times}
\usepackage{latexsym}
\usepackage[dvips]{graphicx}
\usepackage{algorithmic}
\usepackage{algorithm}
\usepackage{eqparbox}
\usepackage{framed}
\usepackage{color}
\usepackage[small]{caption}
\usepackage{subcaption}

\emnlpfinalcopy

\def\emnlppaperid{***}

\title{Keyphrase Extraction using Sequential Labeling}

\author{Sujatha Das Gollapalli \and Xiao-li Li\\
Institute for Infocomm Research, A*STAR, Singapore\\
  {\tt \{gollapallis,xlli\}@i2r.a-star.edu.sg}}

\date{}

\begin{document}

\maketitle

\begin{abstract}
\textbf{Keyphrases} efficiently summarize
a document's content and are used in various document
processing and retrieval tasks. Several
unsupervised techniques and classifiers exist
for extracting {keyphrases} from text documents. Most 
of these methods operate at a phrase-level and 
rely on part-of-speech (POS) filters for candidate phrase generation.
In addition, they 
do not directly handle keyphrases of varying lengths. 
We overcome these modeling shortcomings by
addressing keyphrase extraction as a \textit{sequential labeling} task
in this paper. We explore 
a basic set of features commonly used in
NLP tasks as well as predictions
from various unsupervised methods 
to train our taggers. In addition to a more natural modeling for 
the keyphrase extraction problem, we show that tagging models yield 
significant performance benefits over
existing state-of-the-art extraction methods.
% on 
%datasets of scientific research articles.
%In addition to obtaining
%significant performance improvements over existing methods, 
%we show that ``domain knowledge" can be incorporated into
%the training process via simple constraints rather than intricate
%features used in existing methods.
\end{abstract}
\section{Introduction}
\label{sec:intro}
Keyphrases (or keywords) that provide a concise 
representation of documents %provide a concise summary of a document's topics and 
are beneficial to various data mining and web-based tasks~\cite{sigir99jones,mldm05perner,jis10pudota,www10li}.
For this reason, \textit{keyphrase extraction}, the task of automatically
extracting descriptive phrases or concepts from a document 
continues to be studied in text processing research~\cite{acl14hasan}. 

In unsupervised techniques for keyphrase extraction,
individual terms\footnote{\scriptsize We use ``term",``token", and ``word" interchangeably in this paper.}
 in a document are scored using various ``goodness" or ``interestingness"
measures and these term scores are later aggregated to score phrases~\cite{coling10hasan}. For example, the TextRank algorithm builds
a term graph based on neighboring words in a document and computes the score
of each term as the PageRank centrality measure of its corresponding node in the graph~\cite{emnlp04mihalcea}. 

Supervised models use known (``correct")
keyphrases to frame keyphrase identification as a binary classification task. 
Particularly, candidate phrases are assigned positive and negative labels and
features such as the part-of-speech ({POS}) tag sequence of the phrase, TFIDF values~\cite{irbook}, and 
position information in the document are used for training keyphrase classifiers
~\cite{acl14hasan}. Ranking approaches were also 
investigated for keyphrase extraction for specific domains where preference information
among keyphrases is available~\cite{sigir09jiang}.

Candidate phrase generation is a \textit{crucial} step in the keyphrase extraction pipeline
since the extracted phrases form input to the subsequent
scoring or classification modules. A widely-adopted approach
for generating candidates phrases involves first extracting all $n$-grams
from a document (typically, $n=1,2,3$) and retaining $n$-grams 
that satisfy certain POS filters. Several previous works
~\cite{emnlp04mihalcea,aaai08wan,emnlp14caragea} only consider $n$-grams comprising of 
\textbf{nouns} and \textbf{adjectives}; that is, POS tags from the set 
\{NN, NNS, NNP, NNPS, JJ\}.\footnote{\scriptsize The Penn Treebank list of POS tags is available at: https://catalog.ldc.upenn.edu/docs/LDC99T42/tagguid1.pdf.}
This pre-filtering is very effective in discarding $n$-grams unlikely to be keyphrases from
the large set of possible $n$-grams in the document.
For example, single terms marked as prepositions and $n$-grams ending in adjectives
are unlikely to be sensible keyphrases and retaining them adds noise
to the subsequent scoring and classification modules.
\begin{table*}[!bhtp]
\centering
\begin{tabular}{|c|c|c|c|c|c|}
\hline
\textit{Sentence} & \textbf{Keyword} & \textbf{extraction} & for & \textbf{social} & \textbf{snippets} \\
\hline
\textit{L1 (POS) tags} & VB & NN & IN & JJ & NNS \\
\textit{L2 (Phrase) tags} &  VP & NP & PP & NP & NP \\
\hline
\textit{Labels} & KP & KP & O & KP & KP \\
\hline
\end{tabular}
\caption{Example: Tagged title of a research paper in our dataset.}
\label{tab:anec}
\end{table*}

However, consider as an example, a sentence from our dataset shown in Table~\ref{tab:anec}.
This sentence corresponds to the title of a research paper published in 
the World Wide Web conference in the year $2010$. The keyphrases 
(highlighted in \textbf{bold}) specified by the author
along with their level-1 (POS) and level-2 (phrase) tags 
as identified by the Stanford 
Parser\footnote{\scriptsize http://nlp.stanford.edu:8080/parser/index.jsp} 
are shown in this table. With a filter that only retains phrases containing nouns
and adjectives, we automatically lose the correct candidate phrase
``\textit{Keyword extraction}" in this example (POS tag for ``Keyword" is VB).
Indeed, we observed that several author-specified keyphrases in our datasets have POS tags other 
than nouns and adjectives
possibly due to erroneous POS tagging~\cite{acl97samuelsson,cicling11manning} 
or due to an increased range in the types
of phrases/words acceptable to authors.\footnote{\scriptsize ``Google" was not an acceptable \textit{verb} a decade ago.} 

A second drawback in extraction systems that operate at a phrase-level 
pertains to the length of phrases 
considered for scoring or classification. While uni/bi/trigrams 
often suffice for most datasets~\cite{acl14hasan}, it is desirable for a system 
to identify longer phrases when they exist without having to specify 
this as a parameter in the extraction pipeline. %As an example, %though rare~\cite{emnlp14caragea}, 
For example, our datasets (in Section~\ref{sec:expts}) also include several author-specified four-word
keyphrases such as ``software reliability growth model", ``biased minimax probability machine",
and ``web-based personal health record". Once again,
a model that does not consider $4$-gram candidate phrases will exclude 
these correct phrases from the subsequent scoring steps.

\textit{How can we avoid pre-filtering of correct candidate phrases based on
potentially erroneous POS tags during keyphrase extraction? Can we model 
the length of a keyphrase more naturally in our extraction methods?}

In this paper, we answer these 
questions by framing keyphrase extraction as a sequence labeling or tagging task~\cite{kdcd05sarawagi,ftml12sutton}. 
Our tagging task can be defined as follows:
Given a stream of terms
corresponding to the content of a document\footnote{\scriptsize We assume textual content and whitespace tokenization.}, 
assign to each term position, a tag/label from the
set \{KP, O\} where ``KP" corresponds to a keyphrase term and 
``O" refers to a non-keyphrase term. The labels for our example sentence 
are also shown in Table~\ref{tab:anec}. To the best of our knowledge, we are the 
first to systematically address keyphrase extraction as a tagging task 
in order to overcome specific modeling drawbacks in 
existing extraction methods.

\textbf{Our Contributions}: We propose features for learning a keyphrase tagger for textual document content. In contrast with 
existing supervised models that use intricate features 
such as ``the position of the first occurrence of a
phrase divided by the total number of
tokens"~\cite{emnlp03hulth}, ``TFIDF value larger than a threshold"~\cite{emnlp14caragea}, 
we use a basic set of features including term, orthographic, and parse-tree information
for learning our tagger. In addition, we explore the possibility of
adding predictions from unsupervised models as features for training
our tagger. We evaluate our features on publicly-available datasets 
compiled recently from research abstracts of premier conferences in Computer Science.
Our tagger substantially out-performs existing state-of-the-art
keyphrase extraction techniques on these datasets. Additionally, our proposed approach
does not depend on the candidate phrase extraction step thereby 
avoiding problems due to non-inclusion of appropriate phrases during prediction.

We just described, in this section, the motivation for using tagging approaches for keyphrase extraction.
In Section~\ref{sec:methods}, we summarize the features used to train our keyphrase taggers and provide a
brief overview of the state-of-the-art baselines used in comparison experiments. Experimental setup, datasets,
and results are discussed in Section~\ref{sec:expts} followed by an overview of closely-related work
in Section~\ref{sec:related}. Finally, we conclude with directions on future work in
Section~\ref{sec:conclude}.

\section{Proposed Methods}
\label{sec:methods}
Several applications in data mining and NLP involve
predicting an output sequence of labels $\mathbf{y}={<}y_1 \dots y_N{>}$ given 
an input sequence of tokens: $\mathbf{t}={<}t_1 \dots t_N{>}$ where
each input position $i:1 \dots N$ is modeled by vectors of features ${<}\mathbf{x}_1 \dots \mathbf{x}_N{>}$~\cite{kdcd05sarawagi}. 
For example, given a text sequence (of words), Named-Entity Recognition (NER)
involves predicting
labels for each word from the set \{\textit{person, organization, location, other}\} using various dictionary, linguistic,
and surface-pattern features at each position. Taggers involving complex, interdependent
features are often trained using discriminative learning algorithms such as 
Conditional Random Fields (or CRFs). CRFs comprise state-of-the-art models for
several sequence tagging tasks and 
hence we use them for learning a keyphrase tagger~\cite{ftml12sutton}.

\subsection{Features for Keyphrase Tagging}
We use features corresponding to simple surface patterns, terms, parse tags,
and presence of the term in the title of the document for learning our taggers. Our feature types are
listed below:
\begin{enumerate}
\item \textbf{Term, orthographic, and stopword features}: Term features refer to 
the raw tokens corresponding to the textual content 
in the document. We use whitespace tokenization and convert all
tokens to lowercase after removing punctuation. If the term comprises of punctuation only, we
explicitly indicate this using an ``allPunct" feature. We also indicate if the term
is capitalized or corresponds to 
a stopword~\footnote{\scriptsize We used the stopword list from Mallet~\cite{mallet}.}
using the boolean features
``isCapitalized" and ``isStopword" respectively. In addition,
the end of a sentence is explicitly indicated using an ``EOL" feature
to capture the intuition that keyphrase labels do not cross sentence boundaries.
\item \textbf{Parse-tree features}: We obtain the 
lexicalized parse of the document content using the Stanford Parser~\cite{acl05finkel}
 and incorporate the level-1 and level-2 parse
tags for each position as features. These features correspond to the part-of-speech and immediate phrase
tags for a given word. Almost all existing works incorporate \textit{POS} tags in their models due to their
usefulness in identifying keyphrases~\cite{acl14hasan}. Indeed, Hulth~\shortcite{emnlp03hulth} showed that incorporating
linguistic knowledge such as NP-chunking and POS tags 
yields dramatic improvements in keyphrase extraction results over 
using statistical features alone.
\item \textbf{Title features}: We indicate if the term is part of the document's title using a boolean feature (``isInTitle").
The title of a document can be considered a summary sentence describing the document and 
authors often add discriminative terms in their titles to accurately represent their works.
The \textit{isInTitle} feature depends on document structure information that
comprises knowledge of various document sections such as \textit{abstract}, \textit{title}, and \textit{paragraph boundaries} 
and is not always available.
Assuming their availability, title words
were shown to benefit extraction performance in previous studies~\cite{sigir09jiang,semeval10kim}. 
\item \textbf{Unsupervised keyphrase features}: Unsupervised keyphrase extraction 
methods are useful when annotated training data is unavailable to learn supervised models.
These methods designate ``interestingness" of terms using various metrics such as 
``frequency in a corpus" and rank phrases
based on the scores of terms comprising them~\cite{coling10hasan}. 
We obtain the top-$10$ predicted keyphrases from the popular 
unsupervised methods: TFIDF~\cite{coling10hasan},
TextRank~\cite{emnlp04mihalcea}, SingleRank, and ExpandRank~\cite{aaai08wan} and
add boolean features
indicating whether the term is part of the top predicted phrases in each of these methods.
These methods are described further in Section~\ref{subsec:baselines}. 
\end{enumerate} 

We refer to the set of term, orthographic, stopword, and parse-tree features as the \textbf{Basic} set in our experiments.
These features are fairly standard in NLP tagging tasks~\cite{book10indurkhya}. 
For keyphrase extraction, specialized features based on the title (\textbf{Title}) as well as 
those based on unsupervised extraction methods (\textbf{UKP}) improve performance beyond
that afforded by the ``Basic" features.

Note that our features are fairly simple in design compared to 
some intricate features used in existing state-of-the-art 
supervised models. Example such features include: ``the position of the first occurrence of a
phrase divided by the total number of tokens"~\cite{ijcai99frank,emnlp03hulth}, ``the distribution
of terms among different document sections"~\cite{icadl07nguyen}, ``the distance of the first occurrence of a
phrase from the beginning of a paper is below some value $\beta$"~\cite{emnlp14caragea}.

\textbf{Incorporating information from neighboring tokens}: Given the token stream corresponding to the document content,
let F,G represent feature-types described above (term, POS etc.) and $i$ 
represent a token position. The feature templates used for
training our keyphrase tagger are listed in Table~\ref{tab:ftemplates}.
\begin{table}[htp]
\centering
\begin{tabular}{ll}
\hline
Unigram features & $F_i$, $F_{i-1}$, $F_{i+1}$ \\
Bigram features & $F_{i-1}F_{i}$ and $F_{i}F_{i+1}$ \\
Skipgram features & $F_{i-1}F_{i+1}$ \\
Compound features & $F_{i}G_{i}$ \\
\hline
\end{tabular}
\caption{Feature templates for the CRF tagger.}
\label{tab:ftemplates}
\end{table}

In the above template, the unigram features refer to the features
generated at position $i$ using the token at that position. That is, the
actual term, POS tag, capitalization and other features refer to the token 
at the exact position $i$
in the token stream. The neighborhood information for a given position is incorporated using
the bigram and skipgram features that reference tokens at the previous
and next positions relative to $i$. Intuitively, if a current token is part of
a multiterm phrase, this may be indicated via suitable bigram and skipgram features (e.g.,
they may share a particular POS tag sequence or have the same phrase tags). Compound features are conjunctive features 
indicating stronger hints to the tagger. For example, the combined feature ``isInTitle" and
``part of TFIDF extracted phrase" is likely to indicate stronger evidence to the tagger than 
each of these features in isolation. 

\begin{table*}[!thp]
\centering
\begin{small}
\begin{tabular}{|l|l|}
\hline
\textbf{Type of feature} & \textbf{List of features} \\
\hline
Basic & social (term), L1-JJ (POS tag), L2-NP (phrase tag), noCap, notStop \\
Title and UKP & isInTitle, TFIDF, TR, SR, ER, allUKP, AtleastOneUKP, AtleastTwoUKP \\
Bigrams & BIG1-social\_snippets, BIG1-L1-JJ\_L1-NNS, BIG-1-isStopword\_notStop, BIG-1-NoUKP\_TFIDF \\
Skipgrams & SKIP-1-for\_snippets, SKIP-1-L1-IN\_L1-NNS, SKIP-1-L2-PP\_L2-NP SKIP-1-notInTitle\_isInTitle\\
Compounds & CMPD-L1-JJ\_L2-NP, CMPD-L2-NP\_isInTitle, CMPD-L2-NP\_noCap, CMPD-notStop\_TFIDF \\
\hline
\end{tabular}
\end{small}
\caption{\small Sample features are shown for the token ``social" in the example from Table~\ref{tab:anec}.}
\label{tab:anecfeats}
\end{table*}
\textbf{Illustrative Example}: We provide a partial list of features extracted 
for the term ``social" from our anecdotal example sentence
in Table~\ref{tab:anecfeats} as illustration where the phrase ``social snippets" was correctly
identified as a keyphrase by the unsupervised keyphrase extraction methods: TFIDF, TextRank,
SingleRank and ExpandRank. This match is indicated by the features listed as 
TFIDF, TR, SR, and ER in Table~\ref{tab:anecfeats}. In addition, the basic features comprising of the 
POS and phrase tags, lack of capitalization (noCap) and no match in the stopwords list (notStop)
are shown. 

The ``BIG1" and ``BIG-1" prefixes indicate bigrams involving the current token with its next and 
previous tokens respectively. Thus ``BIG-1-isStopword\_notStop" refers to the stopword ``for" 
adjacent to the non-stopword ``social" in the example sentence ``\textit{Keyword extraction for social snippets}".
Similarly, the ``L1-NNS" refers to the POS tag of the next word ``snippets" in the 
feature ``BIG1-L1-JJ\_L1-NNS". In the skipgram ``SKIP-1-for\_snippets", the term features corresponding
the previous and next token with respect to the current position are represented.
In this example, though ``social" is an adjective, it combines with 
a noun to form a noun phrase in the parse tree resulting in the conjunctive feature, ``CMPD-L1-JJ\_L2-NP".

\subsection{Baseline Methods}
\label{subsec:baselines}
\textbf{Classification-based Extraction}: In the previous sections, 
we designed various features for learning tagging models for keyphrase extraction
under the assumption that useful feature and label dependencies exist between adjacent words in text.
For instance, if a term is correctly identified
as a keyphrase token by the tagger, an adjacent term which shares the same phrase label as the current term 
(part of the same phrasal structure in the parse-tree of the sentence), is likely to a keyphrase token too.
CRFs incorporate label dependencies during inference and predict the entire 
label sequence $\mathbf{y}=y_1 \dots y_N$ for a given sequence $\mathbf{t}$ (using notations from Section~\ref{sec:methods}).

However, our features can also be used in classification mode. That is, a label 
can be predicted for each position independently for $i: 1 \dots N$. In classification,
the pairs $(\mathbf{x}_i, y_i)$ form training examples rather 
than $({<}\mathbf{x}_1 \dots \mathbf{x}_N{>}, {<}y_1 \dots y_N{>})$ forming
a single training example in tagging. We 
evaluate our features using the Maximum Entropy and Na\"ive Bayes classifiers 
in Section~\ref{sec:expts}.

\textbf{KEA and CeKE}: We study two state-of-the-art models for comparisons: (1) the popular KEA system ~\cite{ijcai99frank}
and (2) CeKE, a recent technique designed to incorporate citation information into
the keyphrase extraction process~\cite{emnlp14caragea}.\footnote{\scriptsize Working implementations 
of both systems are available online.} 
While several approaches
exist for keyphrase extraction, most techniques
target specific corpora and rely on features based on document types and domain 
information~\cite{www07bao,sigir08xu,naacl09liu,hlt11zhao,corr13marujo}. 
In contrast, both CeKE and KEA obtain competitive performance using features solely based on document content
and corpus-level information~\cite{acl14hasan}.
We disabled the three citation-network based features in CeKE (referred to as CeKE$^-$ in 
Section~\ref{sec:expts}) in our experiments since these special features
are not commonly available for all documents and we would like to
use features based on document-content and corpus-level properties in all models for a uniform comparison.

Both KEA and CeKE
extract candidate phrases ($n$-grams) and 
use labeled examples to learn binary classifiers
for identifying keyphrases among the qualifying phrases. The KEA system
incorporates features based on TFIDF (a standard measure of term importance 
commonly used in IR) and position information of the phrase in the 
document~\cite{ijcai99frank}. For instance, due to its importance, a keyphrase 
may be expected to appear
early in the document. 
CeKE uses features from KEA as well as additional features such as ``the POS sequence of the phrase" and
``TFIDF larger than a tunable threshold parameter"~\cite{emnlp14caragea}. 

\textbf{Unsupervised Extraction Models}: 
Since predictions from unsupervised keyphrase extraction methods are incorporated as
features within our tagging models, we include them in comparison experiments.
Unsupervised models score terms satisfying stopword and parts-of-speech filters 
based on some criterion and use these scores to 
evaluate phrases or $n$-grams (comprising of consecutive words)~\cite{aaai08wan}. 
For instance, in the TFIDF model, a term's score is the product of its frequency
in the document and its inverse document frequency in the collection~\cite{irbook}. 

The TextRank, SingleRank,
and ExpandRank models construct an undirected graph based on the textual content of the
document~\cite{emnlp04mihalcea,aaai08wan}. The nodes in this graph correspond to non-stopword 
terms in the document satisfying POS filters and the edges indicate the neighborhood of terms. 
In TextRank, this neighborhood is restricted to adjacent terms and edges are assigned uniform weights
whereas in SingleRank and ExpandRank, the terms occurring within a window
(parameter to the model) can be connected by edges in the graph. Both SingleRank
and ExpandRank compute weights for edges using term frequency information within
the document. ExpandRank additionally uses information from textually similar documents
in assigning edge weights. All the three models run the PageRank algorithm~\cite{techrep98page} 
on the underlying
term graph and use the corresponding node centrality values as individual term scores.

\section{Experiments}
\label{sec:expts}
\subsection{Datasets}
We employed the datasets collected by Caragea et al.~\shortcite{emnlp14caragea} for illustrating 
the performance of our keyphrase tagger.
To the best of our knowledge, these datasets
comprise the most recent, largest, publicly-available benchmark datasets 
for keyphrase extraction. Research paper abstracts from two
premier research conferences in Computer Science: the World Wide Web (WWW) Conference and 
the ACM SIGKDD Conference on Knowledge Discovery and Data Mining (KDD)
are provided along with author-specified keyphrases in these datasets. These
keyphrases comprise the ``gold standard" for evaluation. That is, we evaluate the keyphrases
extracted by various methods against the author-specified lists of keyphrases
for each paper using the standard evaluation metrics namely, precision, recall, and F1 measure~\cite{irbook}
and average them across all papers in a dataset~\cite{acl14hasan}. Our datasets are summarized
in Table~\ref{tab:ds}. Retaining only the keyphrases
that occur in the associated abstracts,
we have, on average, about $150$-$170$ terms in each abstract with about $2.32$ keyphrases ($4$-$5$ terms)
per abstract.\footnote{\scriptsize Occasionally, the author-specified keyphrase
is not present \textit{as is} in the document content. For example, the author might say ``CRF" 
in the keyphrase list but use ``Conditional Random Fields" in the rest of the document.}
\begin{table}[!htp]
\hspace*{-0.5cm}
\centering
\begin{small}
\begin{tabular}{|c|c|c|c|}
\hline
\textbf{Venue} & \textbf{\#Abstracts} & \textbf{\#Total KPs/Terms} & \textbf{\#Total Terms}\\
\hline
WWW & 588 & 1365 (2293) & 88853 \\
KDD & 263 & 611 (1116)& 44955\\ %  2.32	\\ %611
\hline
\end{tabular}
\end{small}
\caption{Summary of Datasets} 
\label{tab:ds}
\end{table}
\begin{figure*}[!htp]
\begin{minipage}[b]{0.45\linewidth}
\centering
\includegraphics[scale=0.55]{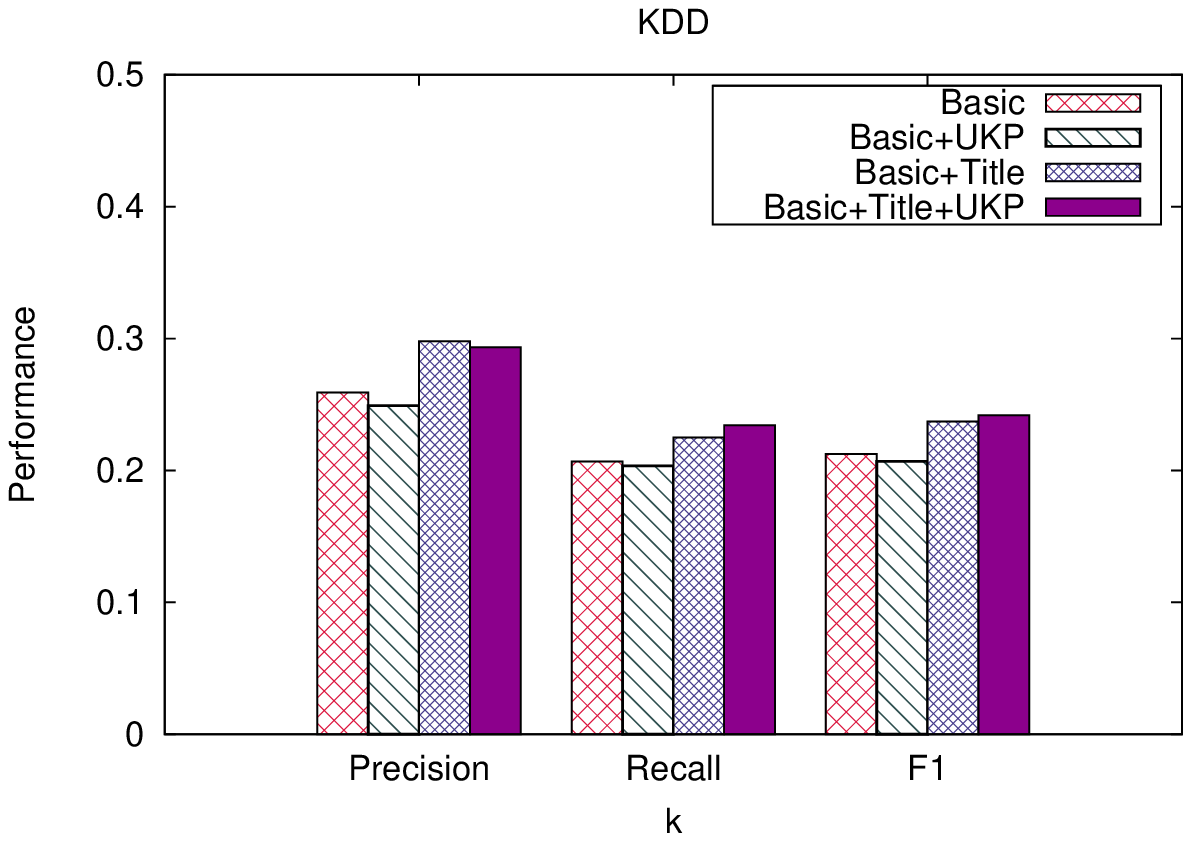}
\end{minipage}
\hspace{0.45cm}
\begin{minipage}[b]{0.45\linewidth}
\centering
\includegraphics[scale=0.55]{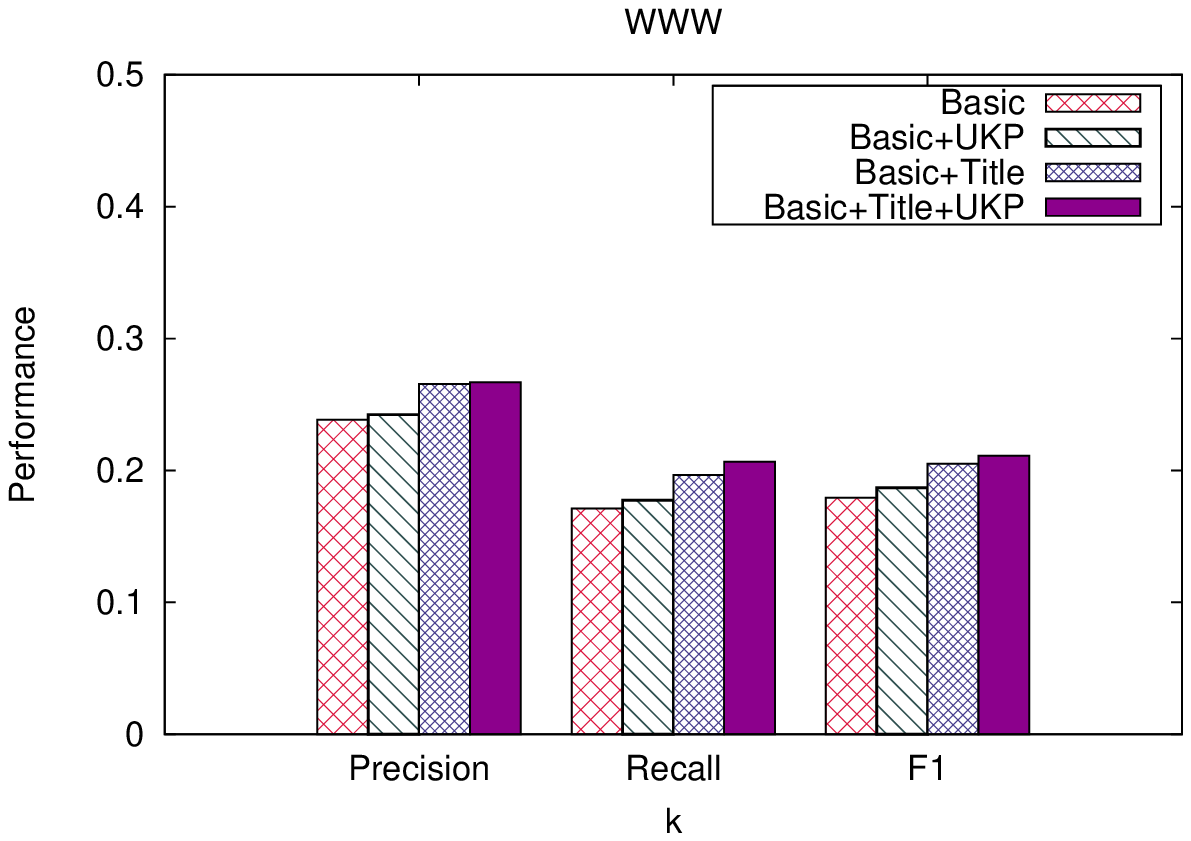}
\end{minipage}
\caption{Effect of various types of features}
\vspace{-10pt}
\label{fig:featsets}
\end{figure*}

We used the CRF implementation from Mallet~\cite{mallet}
and the publicly-available implementations for KEA\footnote{\scriptsize http://www.nzdl.org/Kea/} and 
CeKE\footnote{\scriptsize http://www.cse.unt.edu/$\sim$ccaragea/keyphrases.html} 
in our experiments.\footnote{\scriptsize All code and processed data will be made publicly available.} 
We also used the Maximum Entropy and Na\"ive Bayes classifier implementations from Mallet for testing
our features in the classification mode. 
All reported numbers are averages of five-fold cross validation experiments.
 
\subsection{Results and Discussion}
\textbf{\textit{Tagging performance}}: Tables~\ref{tab:wwwperf} and~\ref{tab:kddperf} summarize the performance 
of our taggers compared to KEA and CeKE$^-$ as well as MaxEnt (Maximum Entropy) and Na\"ive Bayes models
trained using our features. In the
classification mode, ``KP" or ``O" labels are predicted for each term position 
independently and consecutive ``KP" terms are extracted as a keyphrase.
This setting is in contrast to both CeKE and KEA which first extract candidate phrases, compute features
and later classify them.

\label{subsec:results}
\begin{table}[htp]
\begin{small}
\begin{center}
\begin{tabular}{|lccc|}
\hline
\textbf{Method} & \textbf{Precision} & \textbf{Recall} & \textbf{F1} \\
\hline
CRF & \textbf{0.2669} & 0.2067& \textbf{0.2112} \\
%%%%%CRF+PR & 0.2365& 0.3398& 0.2553 \\
MaxEnt & 0.1702 & 0.1562 & 0.1479 \\
Na\"ive Bayes & 0.0808 & \textbf{0.4851} & 0.1344 \\
\hline
CeKE$^{-}$ & 0.1715 & 0.2038& 0.1585 \\
KEA & 0.1228 & 0.2841& 0.1626 \\
\hline
\end{tabular}
\end{center}
\end{small}
\caption{Five-fold CV performance on WWW}
\label{tab:wwwperf}
\end{table}

\begin{table}[htp]
\begin{small}
\begin{center}
\begin{tabular}{|lccc|}
\hline
\textbf{Method} & \textbf{Precision} & \textbf{Recall} & \textbf{F1} \\
\hline
CRF & \textbf{0.2933} & 0.2343 & \textbf{0.2417} \\
%%%%CRF+PR & 0.2181 & 0.4606 & 0.2768 \\
MaxEnt & 0.1682 & 0.1753 & 0.1551 \\
NaiveBayes & 0.0773 & \textbf{0.5262} & 0.1312 \\
\hline
CeKE$^{-}$ & 0.2104 & 0.2673 & 0.2010 \\
KEA & 0.1589 & 0.3716 & 0.2133 \\
\hline
\end{tabular}
\end{center}
\end{small}
\caption{Five-fold CV performance on KDD}
\label{tab:kddperf}
\end{table}
\begin{figure*}[!bthp]
\begin{minipage}[b]{0.45\linewidth}
\centering
\includegraphics[scale=0.55]{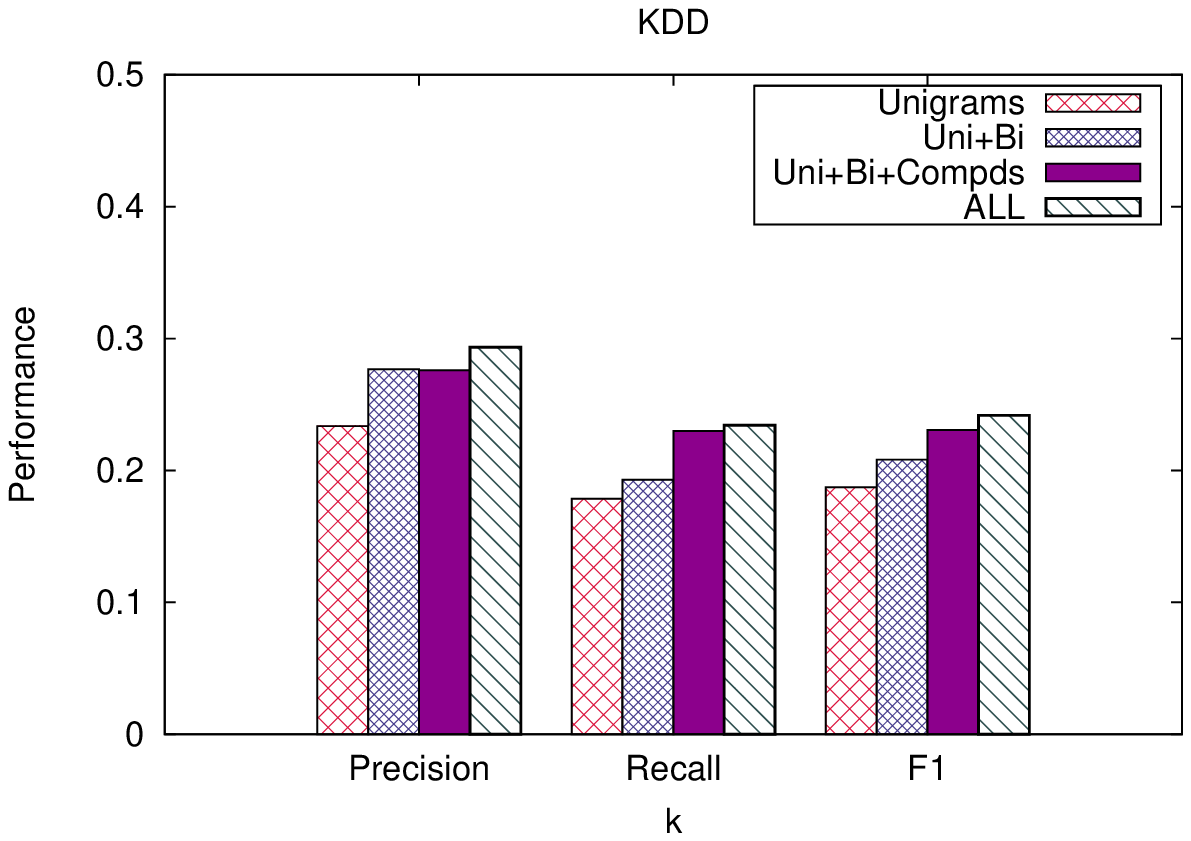}
\end{minipage}
\hspace{0.45cm}
\begin{minipage}[b]{0.45\linewidth}
\centering
\includegraphics[scale=0.55]{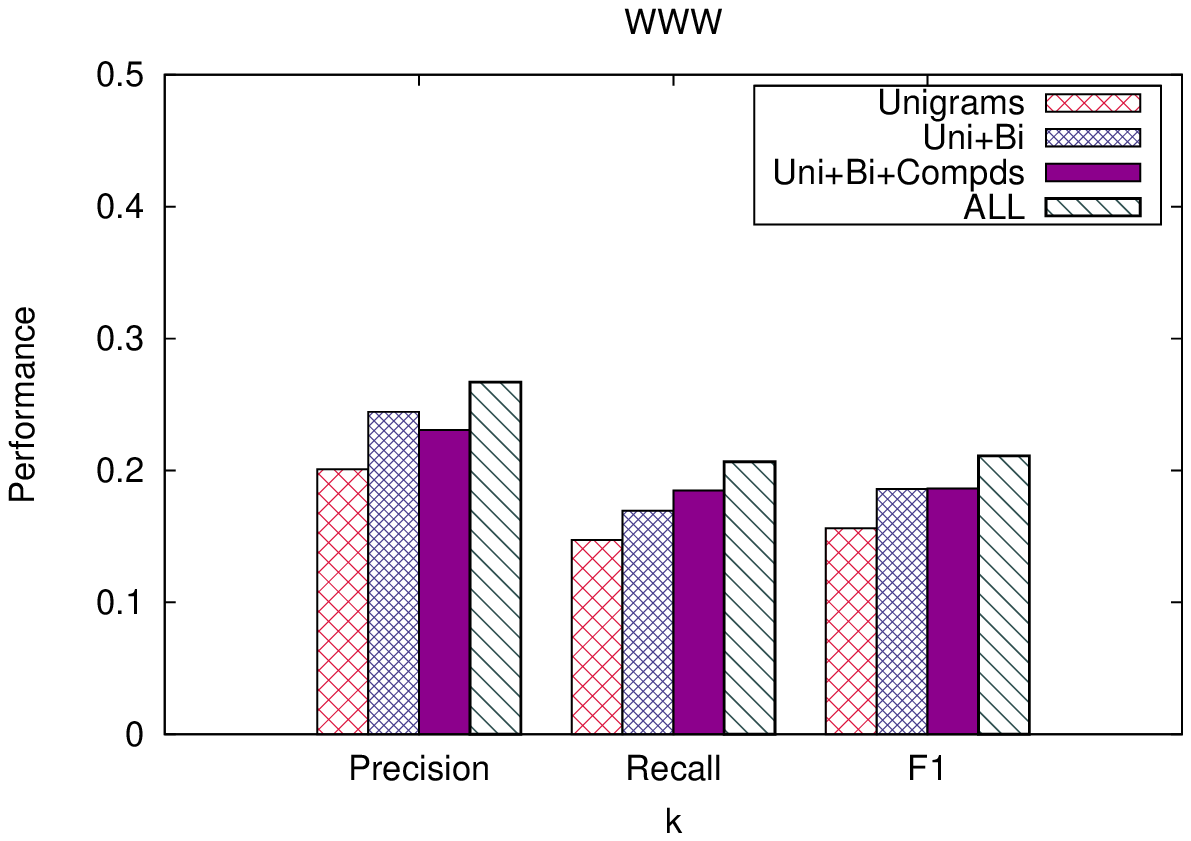}
\end{minipage}
\caption{Performance with Bigram/Skipgram/Compound Features}
\vspace{-10pt}
\label{fig:l1l2}
\end{figure*}

As highlighted in Tables~\ref{tab:wwwperf} and~\ref{tab:kddperf}, our tagger
performs significantly
better than classification approaches despite using relatively simpler features.
The precision of extraction with our CRF taggers is superior to
both the term-level (MaxEnt and NaiveBayes)
and phrase-level (CeKe$^{-}$ and KEA) classifiers, and the overall F1 scores are significantly higher than the other
competing models. Both KEA and the Na\"ive Bayes classifier trained on our features obtain good recall 
but at the cost of low precision. Despite
using the same features, the performance is significantly
lower with MaxEnt and Na\"ive Bayes models in Tables~\ref{tab:wwwperf} and~\ref{tab:kddperf}. These
numbers confirm that models like CRFs
that incorporate feature-label combinations during inference
are better suited for this task than term-level classifiers.

\textbf{\textit{Effect of different feature types}}:The plots in Figure~\ref{fig:featsets} show
the performance of our taggers with different feature sets on both our datasets.
Though the ``Basic" feature set does reasonably well, the performance can be 
significantly enhanced by augmenting the ``Basic" features with 
``Title" features but only slightly with the ``UKP" features.

As witnessed by the improved precision, title terms are particularly
important in scientific documents and often form parts of the 
correct keyphrases. Adding features based on unsupervised predictions
seems to improve the recall slightly resulting in an overall boost in 
the F1 measure using the three sets of features. 

\textbf{\textit{Effect of Bigrams/Skipgrams/Compound features}}:
In Section~\ref{sec:methods}, we described
bigrams, skipgrams and compound features for capturing the neighborhood 
information at a given term position. In Figure~\ref{fig:l1l2}, we show the effect of 
each of these feature templates on our datasets. The
increasing F1 values with the addition of each set of features
substantiates the use of neighborhood features in predicting 
the correct label for a given token. However, these features need to be 
coupled with labels during training and inference as handled in CRFs since they
are not very effective in classifiers (Tables~\ref{tab:wwwperf},~\ref{tab:kddperf}). 

\begin{figure*}[!bhtp]
\begin{minipage}[b]{0.45\linewidth}
\centering
\includegraphics[scale=0.55]{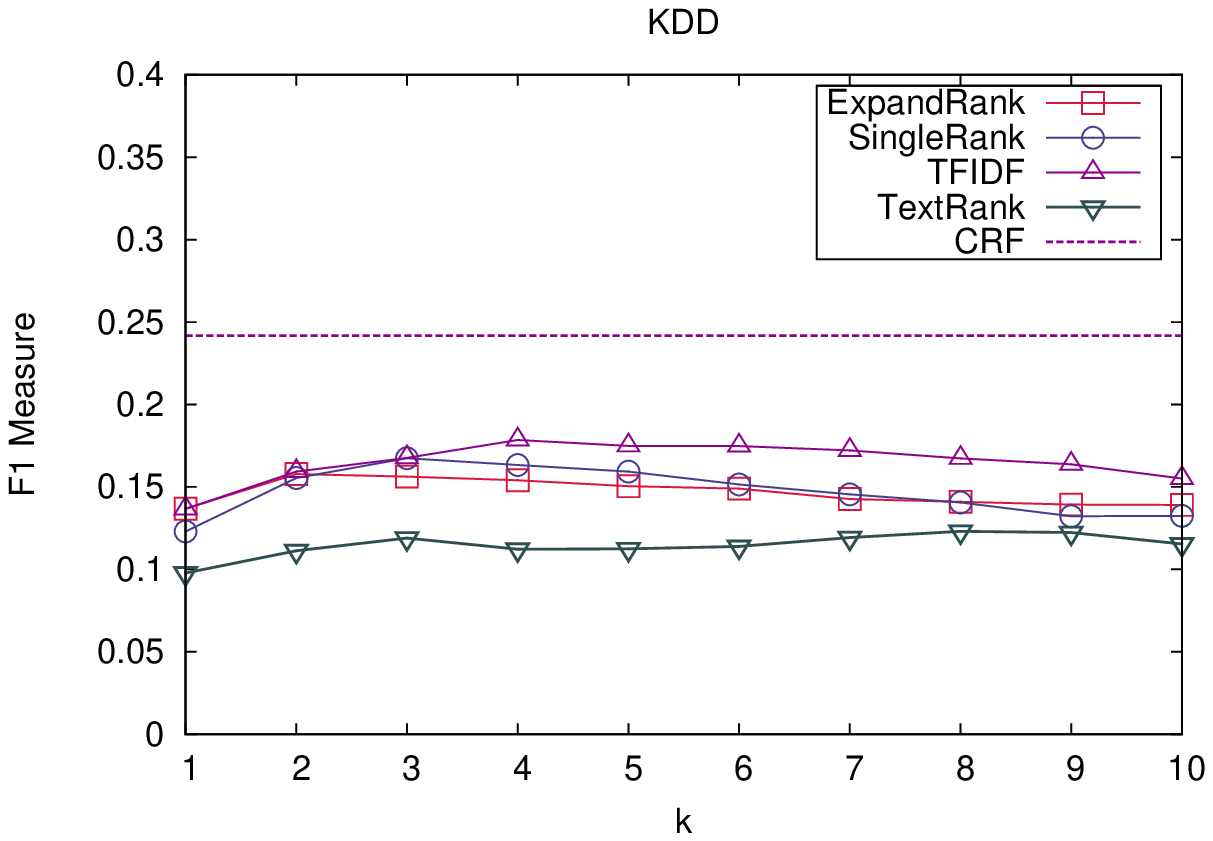}
\end{minipage}
\hspace{0.45cm}
\begin{minipage}[b]{0.45\linewidth}
\centering
\includegraphics[scale=0.55]{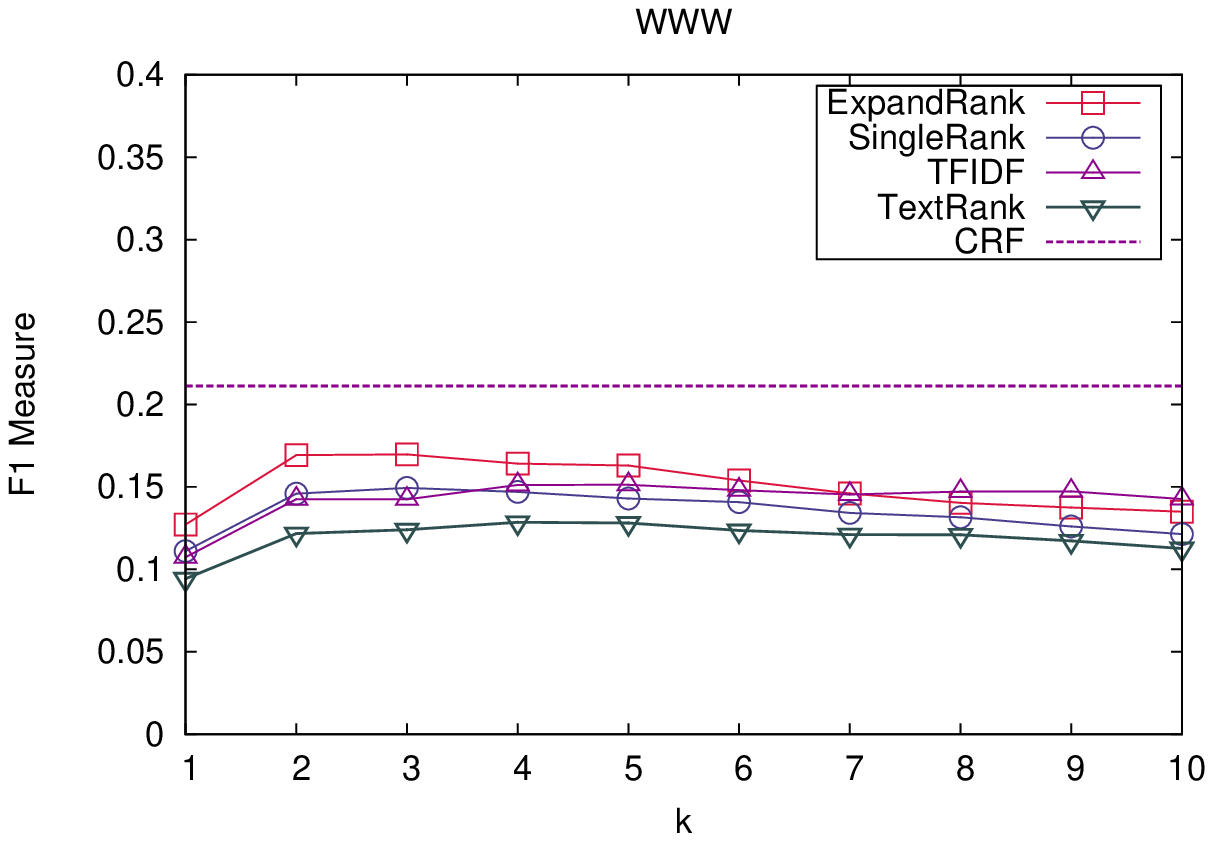}
\end{minipage}
\caption{F1 plots with Unsupervised Keyphrase Extraction Techniques.}
\vspace{-10pt}
\label{fig:ukpcomp}
\end{figure*}

\textbf{\textit{Comparison with unsupervised methods}}:
In the plots of Figure~\ref{fig:ukpcomp}, we show the F1 plots for
the different unsupervised keyphrase extraction techniques listed 
in Section~\ref{subsec:baselines} and the F1 obtained with our CRF tagger.
Evaluation metrics for unsupervised methods are usually 
presented on top-$k$ predictions since all candidate phrases are assigned scores in 
these methods~\cite{coling10hasan}. The F1 values obtained on both the datasets with 
all the unsupervised techniques 
investigated are considerably lower than those obtained with our taggers. 

\begin{table}[!htp]
\begin{small}
\begin{center}
\begin{tabular}{|lccc|}
\hline
\textbf{Train/Test} & \textbf{Precision} & \textbf{Recall} & \textbf{F1} \\
\hline
WWW/KDD & 0.2551& 0.1921 & 0.2012 \\
%%%&CRF+PR & 0.2715 & 0.2621 & 0.2455 \\
\hline
KDD/WWW & 0.1956& 0.1553 & 0.1583 \\
%%%& CRF+PR & 0.2176& 0.2017& 0.1883 \\
\hline
\end{tabular}
\end{center}
\end{small}
\caption{Performance across datasets}
\label{tab:crossperf}
\end{table}

\textbf{Performance across datasets}: Finally, we demonstrate the 
transferable nature of our proposed features in Table~\ref{tab:crossperf}. 
The model trained on the WWW dataset is used to 
predict keyphrases for the KDD dataset and vice-versa in this experiment. 
As shown by the numbers in this table, our models continue to be reasonably effective even 
when trained on a different (though related) collection of documents. 
The tagging performance is better with WWW (as the training dataset)
possibly due its larger size.

Based on the experiments describe in this section, we conclude that
a tagging-based approach not only overcomes several modeling shortcomings
in existing state-of-the-art keyphrase extraction models, but also attains
significant improvements in the extraction performance. Additionally, unlike
existing supervised approaches, they are trained using relatively simpler features.

\section{Related Work}
\label{sec:related}
Keyphrase extraction is a widely-studied task in various
domains~\cite{ijcai99frank,semeval10kim,dtmbio11bong} 
as well as on different document-types~\cite{naacl09liu,hlt11zhao,corr13marujo}.
Tag recommendation, that also involves keyphrase extraction, is
often studied in web contexts~\cite{www07bao,sigir08xu}. Several supervised
and unsupervised methods exist for keyphrase extraction. We refer
the interested readers to recent survey articles by Hasan et al.~\shortcite{coling10hasan,acl14hasan}
for these discussions
and summarize works closest to our approach in this section.

Supervised techniques make use of specialized features such as POS tags, position of the word,
TFIDF values and other features specific to domains and document-types to train keyphrase classifiers
~\cite{ijcai99frank,dl99witten,ir00turney,emnlp03hulth}. 
For example, Caragea at al. designed features using citation contexts to enhance
keyphrase extraction in scientific documents~\shortcite{emnlp14caragea}. 
In contrast, unsupervised approaches characterize the
``goodness" or ``interestingness" of terms based on measures such as 
inverse document frequencies, topic proportions, and graph centrality measures
and used these measures to score phrases in documents~\cite{emnlp04mihalcea,emnlp10liu,ijcnlp13boudin,aaai14gollapalli}. 
\\\\
Tagging approaches and specifically, Conditional Random Fields (CRFs)
were briefly investigated for keyphrase extraction in previous studies.
Bhaskar et al.~\shortcite{coling12bhaskar} study several features
based on document structure such as term presence in various sections like abstract, 
first paragraph, and title as well as
linguistic features such as POS, chunking, and named-entity tags using CRFs for
a small corpus of about 250 scientific articles. Similar features
were employed by Zhang et al. for extracting keyphrases from
documents in Chinese~\shortcite{jcis08zhang}. We complement these preliminary
studies by analyzing why tagging approaches are preferable compared to existing state-of-the-art approaches
and provide an in-depth investigation of keyphrase taggers using relatively larger datasets and simpler features.

\section{Conclusions}
\label{sec:conclude}
We studied keyphrase extraction as a tagging problem 
to overcome specific modeling shortcomings in phrase-level keyphrase extraction techniques. 
Our keyphrase tagger is trained using simple
term, parse, and orthographic features. We also studied the use of 
predictions from existing unsupervised keyphrase extraction models
as features in our tagger. 

Our objective in this paper was to 
illustrate the performance of keyphrase tagging using a basic set of features rather
the intricate features adopted in several state-of-the-art systems. 
However, depending on the domain and the document-type (for example, research papers versus web articles),
external sources of evidence may be available
to enhance keyphrase extraction~\cite{www10li,emnlp14caragea}. 
In future, we would like explore techniques for modeling domain-specific 
aspects as simple features and constraints 
within the tagger for further improving its extraction performance.
\vspace{-10pt}
\bibliography{main}
\bibliographystyle{emnlp2016}

\end{document}